%% file: main.tex
\PassOptionsToPackage{hidelinks}{hyperref}
\pdfoutput=1
\documentclass[electronics,article,submit,moreauthors,pdftex]{Definitions/mdpi}

\usepackage{amsmath,amssymb}
\usepackage{siunitx}
\usepackage{subcaption}
\graphicspath{{figures/}}

\Title{Perplexity Predicts Protection: Choosing Pretrained Backbones for Worst-Client Fairness in Federated Parameter-Efficient Fine-Tuning}

\Author{Kiran Naseer $^{1,}$*, Samreen Azhar $^{1}$, Umar Shoaib $^{1}$, Haroon Mahmood $^{2}$, Muhammad Awais $^{3}$}

\AuthorNames{Kiran Naseer, Samreen Azhar, Umar Shoaib, Haroon Mahmood, Muhammad Awais}

\address{%
$^{1}$ \quad Department of Computer Science, University of Gujrat, Gujrat, Pakistan\\
$^{2}$ \quad Department of Cybersecurity, College of Engineering, Al Ain University, Al Ain, UAE\\
$^{3}$ \quad Department of Computer Science, College of Computer, Qassim University, Qassim 52571, Saudi Arabia}

\corres{Correspondence: 25016119-003@uog.edu.pk}

\firstpage{1}
\pubvolume{15}
\issuenum{1}
\articlenumber{1}
\pubyear{2026}
\copyrightyear{2026}
\datereceived{}
\daterevised{}
\dateaccepted{}
\datepublished{}
\hreflink{}

\abstract{Federated learning lets multiple parties train a shared model without pooling their data, but a client with far less data than the others can end up poorly served even when the group's average accuracy looks fine. We ask whether the choice of pretrained backbone affects this under LoRA fine-tuning, and whether per-word perplexity on the target text predicts which backbone helps the worst-off client before federated training starts. We ran 313 experiments across three text-classification datasets and three similarly sized backbones (RoBERTa, BERTweet, PubMedBERT), each compared against a task-specific baseline on identical data splits. Lower-perplexity backbones consistently produced larger gains for the worst-performing client, with a rank correlation of $-0.87$ across nine dataset-backbone pairs; a backbone held out of the analysis confirmed the pattern. Personalization with Ditto recovered only 4--12\% of the gap between training alone and full federation, and removing aggregation entirely erased the benefit. A client's update also showed no sign of conflicting with the group's update; the two are close to orthogonal, ruling out one proposed explanation for this failure. Practically: measure perplexity on a sample of task text before choosing a backbone, and do not rely on personalization to protect a data-poor client. We release our code, predictions, and full results for others to test.}


\renewcommand{\linenumbers}{}
\renewcommand{\papercitation}{}
\renewcommand{\journalname}{Preprint}
\doinum{}
\begin{document}

\input{sections/01_introduction}
\input{sections/02_related_work}
\input{sections/03_method}
\input{sections/04_results}
\input{sections/05_discussion}
\input{sections/06_limitations}
\input{sections/07_conclusion}

\authorcontributions{Conceptualization, K.N.; Methodology, K.N.; Software, K.N.; Formal analysis, K.N.; Investigation, K.N.; Data curation, K.N.; Visualization, K.N.; Writing -- original draft, K.N.; Validation, S.A.; Writing -- review and editing, S.A., U.S.; Supervision, U.S. All authors have read and agreed to the published version of the manuscript.}

\funding{This research received no external funding. The APC was funded by Haroon Mahmood and Muhammad Awais.}

\conflictsofinterest{The authors declare no conflict of interest.}

\dataavailability{The code, the run manifest, the committed alignment table with its per-cell predictions, the protocol deviation log, and the complete per-round result log are available at \url{https://github.com/KiranNaseer-AI/fedalign}.}

\acknowledgments{The authors thank the University of Gujrat for computational support.}

\appendixtitles{yes}
\appendix
\input{sections/08_appendix}

\reftitle{References}
\externalbibliography{yes}
\bibliography{refs}

\end{document}

%% file: sections/01_introduction.tex
\section{Introduction}
\label{sec:intro}

Consider a hospital with two hundred labelled records that joins a federation of ten hospitals. The other nine hold tens of thousands each. After training, the shared model reports good average accuracy. Whether it works for the small hospital is a separate question, and it is the question this paper is about.

Average accuracy across clients hides this case by construction. A federation of ten clients can report 90\% average accuracy while one client sits at 30\%, and the client left behind is often the one the federation was built to serve. Small hospitals, minority-language schools and community clinics join federations
precisely because they cannot train alone. Reporting only the mean tells them nothing about what they will receive.

Pretrained models are now the default starting point for federated text classification, and there is good reason to expect them to help. Nguyen et al.~\cite{nguyen2023where} showed that starting from pretrained weights narrows the gap between IID and non-IID federated training, and that client updates from pretrained weights are better aligned with one another than updates from random initialisation. Their comparison is pretrained against random, measured on average accuracy.

Two questions remain open. A practitioner does not choose between a pretrained model and a random one. They choose among several pretrained models, and nothing in the literature says which to pick. Nor does average accuracy answer the question the small hospital is asking. We take up both.

We ran 313 controlled experiments. Three text datasets are crossed with three pretrained backbones, and the backbones are matched in capacity so that pretraining domain is the only systematic difference across the grid: all three are twelve-layer encoders of 110 to 125 million parameters, each pretrained from
scratch on a different corpus. Every cell uses five random partitions. Every foundation model is compared against a task-specific convolutional baseline trained on the identical partition, which makes each comparison paired and removes
the partition draw as a source of noise.

We measure the client that does worst, not the average. Our primary metric scores each client's model on the full global test set, restricted to that client's classes but with out-of-class predictions counted as errors. This matters more than it may appear. A metric computed only on a client's own test split is
maximised by a constant predictor when that client holds a single class, so under extreme skew it rewards a model that has learned nothing.

The protocol, including the hypotheses, the metrics and the analysis code, was fixed before the grid was run. Departures from it are recorded in a deviation log that we release with the code. One pre-registered prediction failed, and we report it as such in Section~\ref{sec:res-prereg}.

\paragraph{Backbone choice can be settled before training starts.}
Ranking candidate backbones by per-word masked-language-model perplexity on the task text reproduces the ranking of their worst-client benefit. This holds in all three datasets, and across the nine grid cells the rank correlation is $-0.867$ ($p = 0.0025$). Because that analysis was exploratory, we committed an interval in advance for a fourth backbone that had not been run, then ran it. The prediction held. Computing the score needs no federated training at all.

\paragraph{Personalization gives up most of the benefit.}
Ditto~\cite{li2021ditto} is designed to improve worst-case client utility, and it did not help here. It recovers between 4 and 12 percent of the distance between purely local training and FedAvg, and a hundredfold change in its regularisation strength moves that by $0.006$ to $0.048$ across the three cells. We read this as a limit set by client data volume rather than a flaw in the method: at our most extreme skew the smallest client holds roughly twenty examples, and personalization has little to work with at that scale.

\paragraph{Aggregation carries the benefit; the frozen features alone do not.}
Removing aggregation destroys the effect. In all five cells we tested this way, the task-specific baseline beat the pretrained model on every single partition.
We also measured the geometry of the adapter updates directly and found no sign of the interference that has been offered as an explanation. The smallest client's update is not being overwritten by the majority.

\subsection{Contributions and Scope}

\begin{enumerate}
\item A backbone selection rule for federated PEFT that requires no federated
      training, validated out-of-sample on a backbone held out of the analysis.
\item Evidence across three domain-matched cells that personalization forfeits
      most of federation's worst-client benefit when clients are data-poor, with
      the regularisation strength swept over two orders of magnitude.
\item Direct measurement of adapter-space geometry and of a no-aggregation
      control, which together rule out two standing explanations for worst-client
      behaviour under extreme skew.
\item A pre-registered protocol with a logged deviation record, including a
      pre-registered prediction that failed, and a public release of the code, the
      run manifest and all per-round results.
\end{enumerate}

%% file: sections/02_related_work.tex
\section{Related work}
\label{sec:related}

\subsection{Federated Learning under Heterogeneous Data}

FedAvg~\cite{mcmahan2017fedavg} set the standard protocol and is still the default baseline. Its weakness is well documented: when clients hold different label distributions, local models drift apart between rounds and the average of those models is worse than any of them would suggest. FedProx adds a proximal term
that keeps each local model near the global one~\cite{li2020fedprox}. SCAFFOLD tracks control variates and subtracts the estimated drift~\cite{karimireddy2020scaffold}.
Both are evaluated on how quickly and how high average accuracy converges.

Label skew is usually simulated with a Dirichlet partition~\cite{hsu2019measuring}, and we follow that convention. A recent survey catalogues the many forms non-IID data can take and the metrics used to report them~\cite{jimenez2024noniid}.
It notes that per-client evaluation is still uncommon, which is the gap our primary metric is designed to close.

\subsection{Parameter-Efficient Fine-Tuning in Federated Settings}

LoRA~\cite{hu2022lora} has become the default way to perform parameter-efficient fine-tuning (PEFT) a large model with few trainable parameters, and it fits federated learning well because only the low-rank factors need to be sent. Aggregating those factors is harder than it looks. Averaging $A$ and $B$ separately is not the same as averaging their product, so the server's update does not match what the clients computed.

Several methods attack this problem from different angles. FFA-LoRA freezes the randomly initialised $A$ matrices and aggregates only $B$, which makes aggregation linear again and halves communication~\cite{sun2024ffalora}. FedSA-LoRA trains
both factors locally but shares only $A$, keeping $B$ on the
client~\cite{guo2025fedsalora}. FLoRA stacks the clients' modules instead of averaging them, which also handles clients that use different ranks~\cite{wang2024flora}. Other work targets heterogeneous client resources~\cite{cho2024heterogeneous, bai2024flexlora} or federated instruction tuning at scale~\cite{zhang2024fedit}. A recent survey covers the area in
full~\cite{yang2025fedlorasurvey}.

These methods differ in how they aggregate, and they are compared on average accuracy and on communication cost. The frozen backbone is treated as given. We are asking a question one level up: given that some pretrained backbone will be used, which one, and what does that choice do to the client that does worst.

\subsection{Pretrained Models and Fairness in Federated Learning}

Nguyen et al.~\cite{nguyen2023where} give the clearest account of why pretraining helps. Starting from pretrained weights narrows the gap between independent-and-identically-distributed (IID) and non-IID federated training, and client updates from pretrained weights point in more similar directions than updates from random initialisation. Their comparison is between
pretrained and random initialisation, measured on average accuracy.

That result frames our question rather than answering it. In practice nobody chooses between a pretrained model and a random one. RoBERTa, BERTweet and PubMedBERT are all pretrained, all available, and all roughly the same size, and the literature offers no guidance on choosing among them for a federated
deployment. Whether the choice matters at all for the worst client, and whether it can be settled cheaply, are open.

Two families of methods address unequal outcomes across clients. The first changes the objective. Agnostic federated learning optimises for the worst mixture of client distributions~\cite{mohri2019agnostic}, and q-FedAvg reweights clients by their loss so that accuracy is spread more evenly~\cite{li2020qfedavg}. The second gives each client its own model. Per-FedAvg uses a meta-learning formulation~\cite{fallah2020perfedavg}, pFedMe uses Moreau
envelopes~\cite{dinh2020pfedme}, and Ditto trains a personal model regularised toward the global one~\cite{li2021ditto}. pFedLoRA carries the same idea into adapter space~\cite{yi2023pfedlora}.

Ditto is the closest of these to our setting and the strongest reported baseline for worst-case client utility, so we take it as the method to beat. Recent surveys map the wider fairness landscape~\cite{salazar2024groupfairness} and a benchmark
has been proposed for client-level fairness in particular~\cite{heilmann2025benchmark}.

One assumption runs through all of this work. These methods were introduced and tested in settings where every client has enough data to train on, so the question of what personalization does to a client with twenty examples has not been asked. Our results suggest the assumption is doing real work.

\subsection{Positioning of This Work}

Table~\ref{tab:related} places our study against the closest work. Three gaps remain open, and we address each.

No study varies \emph{which} pretrained backbone is used, at matched capacity, with the worst client as the primary outcome. Fairness-aware methods have not been evaluated at the level of skew where the smallest client holds tens of examples. The mechanism usually offered for worst-client failure, interference between the minority update and the aggregate, has been inferred from outcomes rather than measured in adapter space.

An earlier conference version of this study looked at one backbone on two datasets with a single random partition, and reported the opposite pattern: DistilBERT+LoRA increased the worst-client gap relative to TextCNN under extreme skew~\cite{naseer2026when}. The present work extends it to a
nine-cell backbone-by-dataset grid, five partitions per cell, a per-client metric that does not degenerate under extreme skew (Section~\ref{sec:metrics}), and matched local computation. Each of these changes independently moves the result in the direction reported here; we do not attempt to apportion credit among them, since the design was not set up to isolate their separate contributions.

\begin{table}[t]
\centering
\caption{Position of this work. \checkmark\ = addressed, P = partial,
$\times$ = not addressed.}
\label{tab:related}
\small
\begin{tabular}{lccccc}
\toprule
 & Fed.\ PEFT & Worst client & Backbone & Mechanism & Multi-seed \\
 &            &              & varied    & measured  &            \\
\midrule
FedProx~\cite{li2020fedprox}            & $\times$ & $\times$ & $\times$ & $\times$ & \checkmark \\
Ditto~\cite{li2021ditto}                & $\times$ & \checkmark & $\times$ & $\times$ & \checkmark \\
Nguyen et al.~\cite{nguyen2023where}    & P & $\times$ & $\times$ & P & \checkmark \\
FFA-LoRA~\cite{sun2024ffalora}          & \checkmark & $\times$ & $\times$ & P & \checkmark \\
FedSA-LoRA~\cite{guo2025fedsalora}      & \checkmark & $\times$ & $\times$ & $\times$ & \checkmark \\
FLoRA~\cite{wang2024flora}              & \checkmark & $\times$ & $\times$ & $\times$ & \checkmark \\
This work                               & \checkmark & \checkmark & \checkmark & \checkmark & \checkmark \\
\bottomrule
\end{tabular}
\end{table}

%% file: sections/03_method.tex
\section{Experimental Design}
\label{sec:method}

\subsection{Backbones, Datasets, and Partitions}
\label{sec:design}

\begin{figure}[t]
\centering
\includegraphics[width=0.78\linewidth]{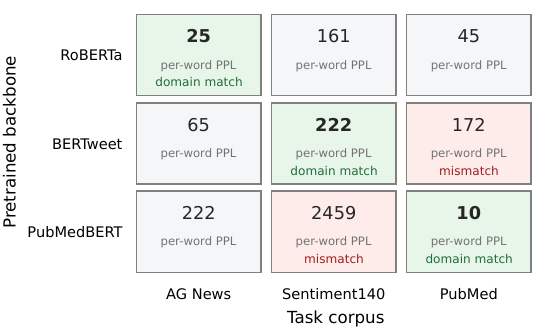}
\caption{The backbone-by-dataset grid. Each cell gives the per-word masked-LM perplexity of that backbone on that corpus. Green marks a domain match, red the two strongest mismatches.}
\label{fig:design}
\end{figure}

The study crosses three text datasets with three pretrained backbones, giving nine cells. Figure~\ref{fig:design} shows the grid.

The backbones are matched in capacity on purpose. All three are twelve-layer encoders of 110 to 125 million parameters, and each was pretrained from scratch on a different corpus. If we had compared a small general-domain model against a large domain-specific one, any difference we found could be attributed to size
rather than to pretraining domain. Matching capacity removes that explanation. Table~\ref{tab:grid} lists the three.

\begin{table}[t]
\centering
\caption{The three backbones. All are base-scale twelve-layer encoders trained
from scratch, so pretraining corpus is the only systematic difference across the
grid.}
\label{tab:grid}
\begin{tabular}{llrl}
\toprule
Backbone & Reference & Params & Pretraining corpus \\
\midrule
RoBERTa-base & \cite{liu2019roberta}     & 125M & General web text \\
BERTweet     & \cite{nguyen2020bertweet} & 125M & 850M English tweets \\
PubMedBERT   & \cite{gu2021pubmedbert}   & 110M & PubMed abstracts \\
\bottomrule
\end{tabular}
\end{table}

Three cells pair a backbone with the corpus closest to its own pretraining domain.
We call these the domain-matched cells. Two cells pair a backbone with the corpus furthest from it, BERTweet on biomedical abstracts and PubMedBERT on tweets. The remaining four fall in between. Predictions for every cell were written down before any federated run, as described in Section~\ref{sec:alignment}.

Each cell is compared against a task-specific baseline, TextCNN~\cite{kim2014textcnn}, with 2.7M parameters. We use the same tokenizer for the baseline across all three datasets so that tokenisation does not vary with the cell.

We use three English classification corpora that differ in domain and in the number of classes. AG News is news topic classification with four
classes~\cite{zhang2015character}. Sentiment140 is binary tweet
sentiment~\cite{go2009twitter}. PubMed 20k RCT is five-way sentence-role
classification drawn from the abstracts of randomised controlled
trials~\cite{dernoncourt2017pubmed}.

We subsample every dataset to 20{,}000 training examples with a fixed seed. The three corpora differ in size by two orders of magnitude, so without subsampling, dataset size would vary together with domain and a difference between cells could not be attributed to either one. Section~\ref{sec:res-controls} reports a control at the full 120{,}000
examples for one cell. Sequences are truncated at 128 tokens throughout.

Clients are formed with a label-Dirichlet partition~\cite{hsu2019measuring} at
concentration $\alpha$, with $K = 10$ clients. Smaller $\alpha$ concentrates each
class on fewer clients. At $\alpha = 0.1$ the smallest client typically holds
around twenty training examples while the largest holds several thousand.

Partition indices are generated once per (dataset, $\alpha$, seed) and written to
disk. Every model that runs at that seed reads the same file. A foundation model
and its TextCNN baseline therefore see the identical clients holding the identical
examples, which makes each model comparison paired and removes the partition draw as a source of variance — not a minor concern, since on one cell the worst-client score varies by more than 0.25 across partition draws, larger than most of the effects we report.

Each client's shard is split 80/10/10 into training, validation and test data.
For PubMed the split is made at the level of the abstract rather than the
sentence. Sentences from one abstract share topic, style and often vocabulary, so
a sentence-level split would place near-duplicates in both the training and test
halves of the same client and inflate every number we report. The dataset provides
an abstract identifier, and our 20{,}000-sentence subsample spans 11{,}177
abstracts. For the other two datasets no such grouping is needed and the split is
stratified by label.

\subsection{Models, Adaptation, and Federated Protocol}

Each backbone is used with its own tokenizer and kept frozen. We insert LoRA adapters~\cite{hu2022lora} at the query and value projections of every attention sublayer, with rank $r = 8$, scaling $\alpha_{\text{LoRA}} = 32$ and dropout 0.1. A linear classification head is trained alongside the adapters. Only the adapters and the head are trained and communicated.

We implement LoRA directly rather than through a library. Two reasons. The effective weight update $\Delta W = \tfrac{\alpha}{r} BA$ can then be extracted exactly for the geometry measurements in Section~\ref{sec:res-mechanism}, and the
FFA-LoRA variant, which freezes $A$, becomes a single flag rather than a separate code path.

Table~\ref{tab:config} gives the configuration. Rounds are uniform across $\alpha$
so that convergence budget does not vary with the condition. Local epochs are
equal for both model classes, which removes a confound present in much of the
literature: a foundation model given five local epochs against a baseline given
one is not a controlled comparison. Section~\ref{sec:res-controls} reports a
stronger version of the same control, in which the number of local gradient steps
rather than epochs is equalised.

\begin{table}[t]
\centering
\caption{Federated configuration. All conditions share these settings unless a
control states otherwise.}
\label{tab:config}
\begin{tabular}{ll}
\toprule
Parameter & Value \\
\midrule
Clients $K$              & 10 \\
Rounds                   & 30, uniform across $\alpha$ \\
Local epochs $E$         & 1 for both model classes \\
Batch size               & 32 \\
Optimiser                & AdamW \\
Dirichlet $\alpha$       & $\{0.1, 0.3, 1.0\}$ \\
Seeds per cell           & 5 \\
LoRA                     & $r=8$, $\alpha_{\text{LoRA}}=32$, dropout 0.1, $Q$ and $V$ \\
Max sequence length      & 128 \\
Training examples        & 20{,}000 per dataset \\
\bottomrule
\end{tabular}
\end{table}

We compare five aggregation strategies. FedAvg~\cite{mcmahan2017fedavg} is the
default. FedProx~\cite{li2020fedprox} adds a proximal term with $\mu = 0.01$.
FFA-LoRA~\cite{sun2024ffalora} freezes the $A$ matrices and aggregates only $B$.
Ditto~\cite{li2021ditto} trains a personal model per client, regularised toward
the global model with strength $\lambda$, which we sweep over
$\{0.1, 1, 10\}$. Local-only training removes aggregation entirely and serves as
the lower anchor: it tells us what each client could achieve alone.

\subsection{Metrics and Evaluation}
\label{sec:metrics}

We use \emph{worst-client performance} and \emph{max-min utility}
interchangeably in what follows; both are narrower than algorithmic fairness in general, and we make no broader fairness claim. Reporting the worst client requires a per-client score, and the obvious choices break down under extreme skew. We use three.

\begin{description}
\item[P1, client-local macro-F1.] Macro-F1 on the client's own held-out split.
      This is the utility the client experiences on data like its own.
\item[P2, global balanced accuracy.] Balanced accuracy of the client's model on
      the global test set.
\item[P3, client-class macro-F1.] Macro-F1 restricted to the classes the client
      holds, but computed over the \emph{whole} global test set. This is our
      primary endpoint.
\end{description}

The distinction between P1 and P3 decides what the paper measures, so it is worth
setting out. Consider a client that holds a single class, which is common at
$\alpha = 0.1$. Its held-out split contains only that class. A model that predicts
that class for every input scores precision 1 and recall 1 on that split, so its
macro-F1 is 1.0. Under P1 a model that has learned nothing is indistinguishable
from a model that has learned everything. Any metric computed only on the client's
own data has this property, and it is exactly the regime where worst-client
behaviour is interesting.

P3 removes the degeneracy by scoring on the full global test set. The constant
predictor still achieves recall 1 on its class, but it now produces a false
positive on every example from the other classes, so precision falls to roughly
$1/C$ and macro-F1 falls with it. A model must be competent on the client's
classes \emph{and} restrained on the others to score well.

We log all three metrics at every evaluated round, so a reader can recompute any table under either endpoint. Where P1 and P3 disagree we report both and say so, as in Section~\ref{sec:res-ditto}. Across clients we report the minimum, the tenth percentile, mean, harmonic mean and standard deviation. The minimum over ten clients is a noisy order statistic, so the tenth percentile is reported for the primary grid in Table~\ref{tab:A4}.

\subsection{Perplexity Measurement}
\label{sec:alignment}

For every (backbone, dataset) pair we score how well the backbone models the task text, before any federated training. We sample 1{,}500 texts from the training corpus, mask 15\% of tokens at random, and average the masked-language-model loss over three passes~\cite{salazar2020masked}. We also record subword fertility,
which is the mean number of subword tokens per whitespace word, and vocabulary coverage~\cite{rust2021good}.

Token-level perplexity cannot be compared across backbones, because a 64k byte-pair vocabulary and a 30k WordPiece vocabulary segment the same sentence into different numbers of pieces. We convert to per-word perplexity by multiplying the per-token negative log-likelihood by fertility before exponentiating. All
perplexities in this paper are per-word.

We pre-registered a derived quantity rather than the raw score. The protocol specified a two-way centred alignment score, which subtracts the backbone mean and the dataset mean and keeps the interaction. That score, its per-cell values and a signed prediction for every cell were computed and committed to the repository
before the first federated run. Section~\ref{sec:res-prereg} reports what happened
to it.

\subsection{Statistical Analysis and Reproducibility}
\label{sec:stats}

The analysis plan was written and tested on pilot data before the grid was run,
which keeps the choice of test independent of the results it is applied to.

Effects are paired differences taken per seed and tested with a paired $t$-test.
We report exact $p$-values rather than thresholds, Cliff's $\delta$ next to every
test as a distribution-free effect size, and 95\% bootstrap confidence intervals
over seeds with 10{,}000 resamples. Where a table contains several comparisons we
apply Holm--Bonferroni correction within that table.

Five seeds is a deliberate choice with a known cost. The two-sided Wilcoxon
signed-rank test cannot reach $p < 0.05$ at $n = 5$, since its smallest attainable
$p$-value is $0.0625$. We therefore use the paired $t$-test as the primary test
and do not report a nonparametric companion at this sample size. Rank
correlations across cells are Spearman coefficients, and we note the critical
value at the relevant $n$ wherever one is reported.

The protocol, the metrics, the analysis code and the alignment scores with their
signed per-cell predictions were fixed and committed before the grid was launched.
Every subsequent departure is recorded in a deviation log with its date and its
justification, including one pre-registered prediction that failed
(Section~\ref{sec:res-prereg}) and the decision to reduce the number of
conditions when measured compute exceeded our budget.

The full study is 313 completed runs on a single Tesla T4, with no failures. Each
run is a row in a manifest keyed by a hash of its configuration, and results are
appended to a single log as one record per client per evaluated round. We release
the code, the manifest, the committed alignment table, the deviation log and the
complete result log.

%% file: sections/04_results.tex
\section{Results}
\label{sec:results}

Every effect reported below is a paired difference in worst-client macro-F1
between the PEFT model and the TextCNN baseline. At a given seed the two models
train on the same Dirichlet partition, so the difference is taken per seed and
tested paired. A positive value means the pretrained backbone helped the client
that did worst. Unless stated otherwise the metric is P3, the Dirichlet
concentration is $\alpha = 0.1$, and each cell rests on five partitions.

\subsection{Perplexity Predicts Worst-Client Benefit}
\label{sec:res-ppl}

Table~\ref{tab:main} lists all ten cells. Within each dataset we sort the backbones by per-word masked-LM perplexity on that dataset's text. The benefit falls as perplexity rises, and it does so in every dataset. Ordering three backbones correctly by chance has probability $1/6$; three datasets in agreement has probability about $0.005$. This calculation treats the three datasets as independent tests, which they are not: the same three backbones appear in each, so a backbone that is simply stronger overall induces agreement across datasets. We report it as descriptive, for the same reason given for the rank correlation in Section~\ref{sec:limitations}.

\begin{figure}[t]
\centering
\includegraphics[width=0.72\linewidth]{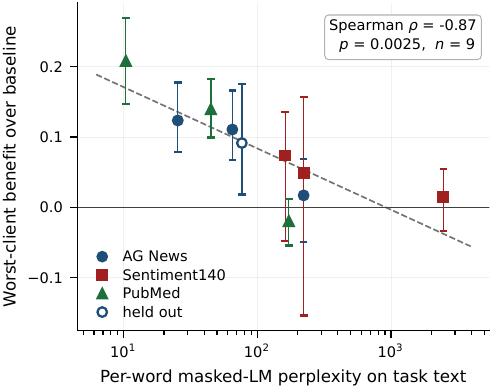}
\caption{Per-word masked-language-model perplexity against worst-client benefit.
Each point is one backbone-dataset pair averaged over five partitions; bars are
95\% bootstrap intervals and marker shape denotes the dataset. The open marker is
DistilBERT, held out of the analysis, whose interval was committed before its runs
executed.}
\label{fig:main}
\end{figure}

Figure~\ref{fig:main} plots the same numbers. Across the nine grid cells the rank
correlation between log per-word perplexity and worst-client benefit is $-0.867$
($p = 0.0025$).

Two cells survive Holm correction within the table. PubMed $\times$ PubMedBERT
gives $+0.208$ ($p = 0.004$, Holm-adjusted $p = 0.037$) and PubMed $\times$
RoBERTa gives $+0.140$ ($p = 0.004$, Holm-adjusted $p = 0.037$). For PubMed
$\times$ PubMedBERT, Cliff's $\delta$ is $1.00$: the foundation model beat the
baseline on every one of the five partitions, with no overlap between the two
sets of scores.

Nine of the ten effects are positive. The exception is PubMed $\times$ BERTweet
at $-0.019$, whose bootstrap interval $[-0.054, +0.012]$ contains zero. We did
not find a cell in which a pretrained backbone clearly harmed the worst client.

The size of the benefit varies with how much room the baseline leaves. On PubMed
the baseline reaches $0.292$ and the best backbone adds $0.208$. On Sentiment140
the baseline already reaches $0.640$ on a binary task, and the largest gain there
is $0.074$. Perplexity orders the backbones within a task. It does not by itself
say how large the gain will be.

\begin{table}[t]
\centering
\caption{All ten cells at $\alpha = 0.1$, grouped by dataset and sorted by
per-word perplexity within each group. Every dataset is monotone. DistilBERT was
held out of the analysis; its interval was committed before its runs executed.
Holm--Bonferroni correction is applied across the ten comparisons in this table.}
\label{tab:main}
\begin{tabular}{llrrrr}
\toprule
Dataset & Backbone & PPL & Effect & $p$ & $\delta$ \\
\midrule
AG News & RoBERTa    & 25.4  & $+0.124$ & 0.013 & $+0.60$ \\
AG News & BERTweet   & 65.0  & $+0.111$ & 0.018 & $+0.52$ \\
AG News & DistilBERT & 76.6  & $+0.091$ & 0.112 & $+0.52$ \\
AG News & PubMedBERT & 221.8 & $+0.017$ & 0.657 & $+0.20$ \\
\midrule
Sentiment140 & RoBERTa    & 161.5  & $+0.074$ & 0.291 & $+0.60$ \\
Sentiment140 & BERTweet   & 222.3  & $+0.049$ & 0.651 & $+0.60$ \\
Sentiment140 & PubMedBERT & 2458.9 & $+0.015$ & 0.568 & $+0.28$ \\
\midrule
PubMed & PubMedBERT & 10.4  & $+0.208$ & 0.004 & $+1.00$ \\
PubMed & RoBERTa    & 44.9  & $+0.140$ & 0.004 & $+0.76$ \\
PubMed & BERTweet   & 171.9 & $-0.019$ & 0.357 & $-0.20$ \\
\bottomrule
\end{tabular}
\end{table}

\subsubsection{The predictor we pre-registered, and the one that replaced it}
\label{sec:res-prereg}

We pre-registered a different predictor. Our protocol specified a two-way centred
alignment score, which removes backbone quality and corpus difficulty as main
effects and keeps only the interaction between them. The threshold was a Spearman
correlation of at least $0.6$ at $p < 0.05$. We measured $\rho = 0.617$ with
$p = 0.0769$. The threshold was not met.

Inspecting the cells showed why. The centred score marks PubMed $\times$ RoBERTa
as a poor match, because RoBERTa fits news text better than it fits biomedical
abstracts. In absolute terms, though, RoBERTa's per-word perplexity on PubMed is
$44.9$, the second lowest value anywhere in our grid. That cell produced a benefit
of $+0.140$, the second largest we measured. Centring had removed the variable
that carries the signal.

We then examined the uncentred per-word perplexity, which ordered the cells
correctly in all three datasets. This analysis was exploratory, so we treated it
as a hypothesis rather than a result. We chose a backbone that had not been run
and was not used to build the relationship, DistilBERT on AG News. Its per-word
perplexity is $76.6$, which sits between BERTweet at $65.0$ and PubMedBERT at
$221.8$. We recorded a predicted interval of $[+0.017, +0.111]$ in the deviation
log, and only then ran the five partitions. The measured effect was $+0.091$.

DistilBERT is drawn as an open marker in Figure~\ref{fig:main}. Adding it gives a four-point ordering on AG News that is monotone in perplexity: $+0.124$, $+0.111$, $+0.091$, $+0.017$ at perplexities of $25.4$, $65.0$, $76.6$ and $221.8$. We report the failed pre-registration and the out-of-sample test together because neither is interpretable without the other. To be explicit: the perplexity relationship itself was found by inspecting these same nine cells, and is exploratory. The single held-out test on DistilBERT is confirmatory for that one point; it is not a replication of the rule.

\subsection{Personalization Forfeits Most of the Benefit}
\label{sec:res-ditto}

\begin{figure}[t]
\centering
\includegraphics[width=0.95\linewidth]{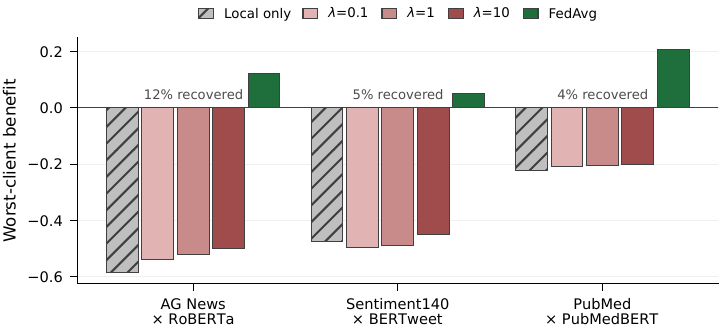}
\caption{Ditto across a hundredfold range of the regularisation strength
$\lambda$, anchored by purely local training and by FedAvg. Every Ditto setting
sits close to the local-only bound.}
\label{fig:ditto}
\end{figure}

Ditto is designed to improve worst-case client utility. In our setting it did not close the gap. Table~\ref{tab:ditto} and Figure~\ref{fig:ditto} show what happened in the three domain-matched cells.

At its default setting Ditto reaches $-0.520$ on AG News, $-0.490$ on Sentiment140 and $-0.203$ on PubMed. Purely local training, with no aggregation at all, reaches $-0.584$, $-0.476$ and $-0.221$ in the same cells. Ditto lands within a few hundredths of the local-only bound every time.

We swept the regularisation strength over two orders of magnitude,
$\lambda \in \{0.1, 1, 10\}$. Larger $\lambda$ pulls each personal model harder
toward the global one, and it does help, but by very little. On AG News the
benefit moves from $-0.539$ to $-0.498$ across a hundredfold change. Expressed as
a fraction of the distance between local training and FedAvg, the best setting of
$\lambda$ recovers 12\% on AG News, 5\% on Sentiment140 and 4\% on PubMed.

A reader may object that as $\lambda$ grows without bound the personal model converges to the global one, so Ditto must eventually match FedAvg. That is true in the limit. What our sweep measures is the rate. A hundredfold increase in $\lambda$ buys between $0.006$ and $0.048$ across the three cells, while the gap left to close is between $0.4$ and $0.7$. Within the range of $\lambda$ used in the literature, personalization gives up nearly all of the protection that federation provides.

We read this as a statement about client data volume, not about the algorithm. At $\alpha = 0.1$ the smallest client holds roughly twenty training examples. A personal model fitted on twenty examples is a poor model, and regularising it toward a good one does not fix that. We cannot fully separate this explanation from others our design does not isolate, including the specific PEFT configuration and the interaction between
local personalisation and the global training objective; data volume is the factor our sweep directly measures, not the only one that could contribute.

\begin{table}[t]
\centering
\caption{Ditto across a hundredfold range of the regularisation strength
$\lambda$, worst-client effect on P3. Every value sits close to the local-only
bound. The final column is the fraction of the local-to-federated gap recovered at
the best $\lambda$.}
\label{tab:ditto}
\begin{tabular}{lrrrrrr}
\toprule
Dataset & Local & $\lambda{=}0.1$ & $\lambda{=}1$ & $\lambda{=}10$ & FedAvg & Rec. \\
\midrule
AG News      & $-0.584$ & $-0.539$ & $-0.520$ & $-0.498$ & $+0.124$ & 12\% \\
Sentiment140 & $-0.476$ & $-0.496$ & $-0.490$ & $-0.448$ & $+0.049$ & 5\%  \\
PubMed       & $-0.221$ & $-0.208$ & $-0.203$ & $-0.202$ & $+0.208$ & 4\%  \\
\bottomrule
\end{tabular}
\end{table}

\paragraph{The result depends on the metric.}
On P3 the recovered fractions are 12\%, 5\% and 4\%. On P1, which scores each
client only on its own held-out split, they are 0\%, $-22\%$ and $+23\%$. The
direction is not stable. P1 is the metric that should favour a personalized model,
and even there Ditto does not consistently beat local training. We state the
result for global competence, which is what a deployed client needs, and we report
the local-split numbers alongside it rather than choosing between them.

\begin{figure}[t]
\centering
\includegraphics[width=0.95\linewidth]{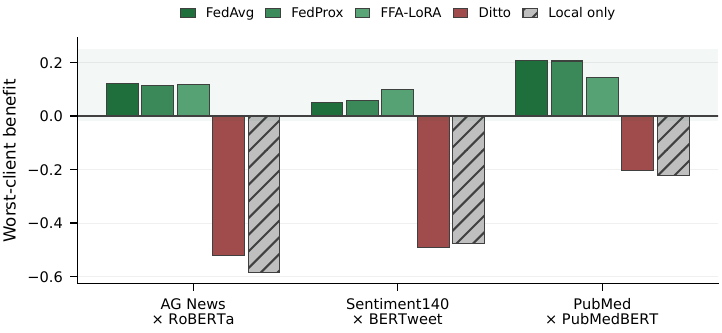}
\caption{Worst-client benefit by aggregation algorithm. The three non-personalised methods cluster above zero; the personalised method collapses onto the local-only bound.}
\label{fig:algos}
\end{figure}

\paragraph{Other algorithms behave differently.}
Figure~\ref{fig:algos} compares five methods in the same three cells. FedAvg,
FedProx and FFA-LoRA cluster together and all stay positive: on PubMed they give
$+0.208$, $+0.206$ and $+0.144$. Ditto and local-only fall well below zero. The
split is not between simple and sophisticated methods. It is between methods that
keep every client on the shared model and methods that do not.

\subsection{Aggregation Carries the Benefit}
\label{sec:res-mechanism}

If a pretrained backbone protects the worst client because its frozen features already separate that client's classes, then removing aggregation should not destroy the effect. We tested this directly with the local-only condition, in which each client trains alone on its own shard.

The effect is large and negative in all five cells we ran: $-0.584$, $-0.476$, $-0.414$, $-0.249$ and $-0.221$. Cliff's $\delta$ is $-1.00$ in every one of them. Table~\ref{tab:A5} gives the cell identity for each value. Without aggregation the baseline beat the pretrained model on every single partition. Whatever the backbone contributes, it does not reach the worst client unless the clients train together.

\begin{figure}[t]
\centering
\includegraphics[width=\linewidth]{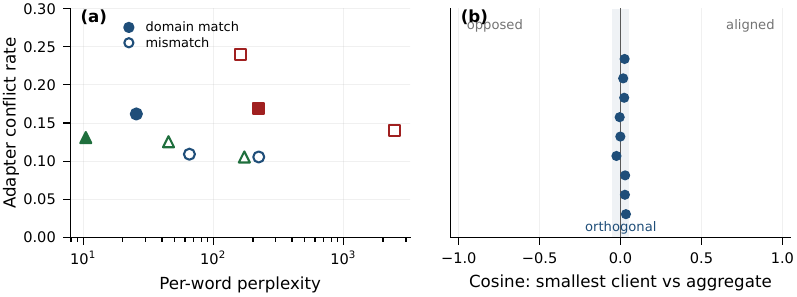}
\caption{Adapter-space geometry. (a) Conflict rate shows no trend with domain
match. (b) The cosine between the smallest client's update and the aggregate is
near zero in every cell, so the two are close to orthogonal rather than opposed.}
\label{fig:mech}
\end{figure}

We also measured the geometry of the adapter updates, because our earlier conference version~\cite{naseer2026when} proposed that the worst client suffers when its update conflicts with the aggregate. Figure~\ref{fig:mech} shows what we found. The cosine between the smallest client's update and the aggregate update averages $+0.013$ over all
cells, with a range from $-0.025$ to $+0.033$. The two are close to orthogonal. They are not opposed. Conflict rates point the same way. Averaged over the last ten rounds, the
domain-matched cells show a conflict rate of $0.154$ and the mismatched cells $0.123$. An interference account predicts the opposite ordering, since conflict should be worst where the backbone fits the task least. We observed no such trend.

Taken together, these two measurements find no evidence for the directional-conflict mechanism examined here. The minority client is not being overwritten by a conflicting majority. Its update
carries little weight in the average, and it gains from the aggregate rather than
losing to it.

\subsection{Controls and Additional Analyses}
\label{sec:res-controls}

\begin{figure}[t]
\centering
\includegraphics[width=\linewidth]{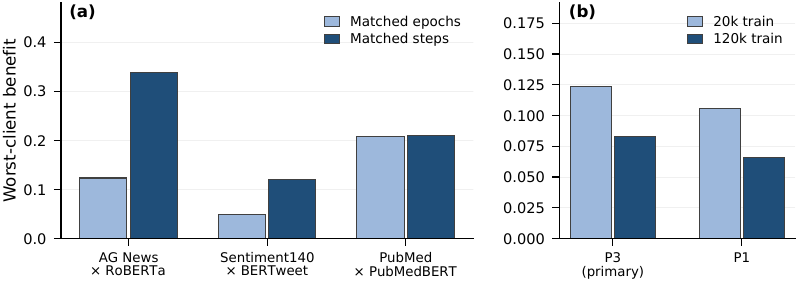}
\caption{Controls. (a) Under matched local gradient steps the benefit grows or
holds in every cell. (b) With six times more training data per client the benefit
falls to about two thirds of its size on both metrics.}
\label{fig:controls}
\end{figure}

\paragraph{Matched local computation.}
The two model classes could in principle differ because one received more local
work per round. We reran the three domain-matched cells with the number of local
gradient steps equalised across both models, giving the baseline its own
matched-steps runs so that the comparison stays paired.
Figure~\ref{fig:controls}(a) shows the outcome. The benefit grows or holds in
every cell: $+0.338$ against $+0.124$ on AG News, $+0.121$ against $+0.049$ on
Sentiment140, and $+0.210$ against $+0.208$ on PubMed. Under a fairer compute
budget the advantage is larger, not smaller. Sentiment140, the weakest cell under matched epochs, more than doubles.

\paragraph{Dataset scale.}
Our main grid subsamples every dataset to 20{,}000 training examples so that
corpus size does not vary with domain. To check that the effect is not an artefact
of that choice, we repeated AG News $\times$ RoBERTa with the full 120{,}000
examples. The benefit fell from $+0.124$ to $+0.083$ on P3 and from $+0.106$ to
$+0.066$ on P1, about two thirds of its original size on both metrics.

The effect survives, but it shrinks. We take this as a boundary condition rather
than a caveat to be minimised: the benefit is largest when the smallest client is
most starved of data, and it fades as every client acquires enough of its own.
That is consistent with the rest of our results.

\begin{figure}[t]
\centering
\includegraphics[width=0.72\linewidth]{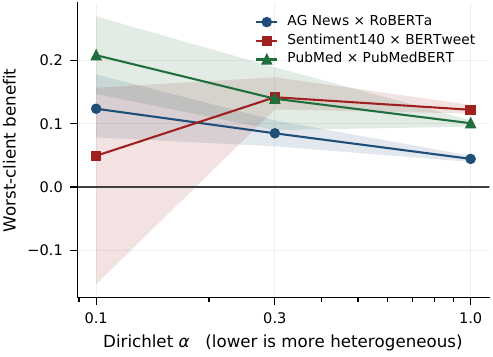}
\caption{Worst-client benefit against the Dirichlet concentration $\alpha$ for the
three domain-matched cells. Bands are 95\% bootstrap intervals over five
partitions.}
\label{fig:alpha}
\end{figure}

\paragraph{Heterogeneity.}
Figure~\ref{fig:alpha} shows the benefit across $\alpha \in \{0.1, 0.3, 1.0\}$.
Two of the three cells strengthen as heterogeneity increases: AG News moves from
$+0.044$ to $+0.124$ and PubMed from $+0.101$ to $+0.208$ as $\alpha$ falls from
$1.0$ to $0.1$. Sentiment140 does not follow the pattern, peaking at $\alpha=0.3$
with $+0.142$ before dropping to $+0.049$. Sentiment140 is a binary task on which
the baseline already scores $0.640$, and it is the same cell that behaves
irregularly in Section~\ref{sec:res-ditto} and under matched compute. We think the
limited headroom explains all three observations, but we have not tested that
directly.

We also checked one cell at $K=50$ instead of $K=10$, keeping
total training examples fixed so each client holds proportionally less data (AG News $\times$ RoBERTa, five seeds). The effect grows sharply: $+0.740$ against $+0.124$ at $K=10$ ($p=0.0001$). TextCNN's worst client fails outright, scoring zero worst-client P3 on two of five seeds, while RoBERTa holds between 0.79 and 0.84 across all five. We read this as consistent with the rest of our results: the benefit is largest when the smallest client is most starved of data, and $K=50$ starves it further still.

%% file: sections/05_discussion.tex
\section{Discussion}
\label{sec:discussion}

\subsection{A Practical Selection Rule}
\label{sec:disc-rule}

The practical form of our main result is a procedure. Before committing to a federated deployment, a practitioner can rank candidate backbones as follows.

\begin{enumerate}
\item Draw a sample of the task text. We used 1{,}500 examples, which was enough for a stable score.
\item For each candidate backbone, mask 15\% of tokens at random and average the masked-language-model loss over three passes.
\item Multiply the per-token loss by subword fertility, the mean number of tokens per whitespace word, before exponentiating. Without this step the scores are not comparable across tokenizers.
\item Rank by the resulting per-word perplexity and take the lowest.
\end{enumerate}

The whole procedure took a small amount of compute per cell on a single Tesla T4. It needs no federated training, no labels and no access to client data beyond a text sample. In our grid it recovered the correct ordering of worst-client benefit in every
dataset, and it placed a fourth backbone correctly on a prediction committed in advance.

Two limits are worth stating plainly. The rule orders backbones within a task; it does not say how large the gain will be. On PubMed the baseline reaches $0.292$ and the best backbone adds $0.208$, while on Sentiment140 the baseline already reaches $0.640$ and the largest gain is $0.074$. Available headroom sets the
scale, and headroom is a property of the task rather than of the backbone. The rule is also relative: a low perplexity tells you which of your candidates fits best, not whether any of them fits well enough to deploy.

The benefit is largest exactly where the stakes are highest. It grows as heterogeneity increases in two of our three cells, and it falls by about a third when every client holds six times more data. Both point the same way. A federation of well-resourced participants with similar data will see little difference
between backbones. A federation in which one participant has a few dozen examples will see a large one.

This is the setting that motivates federated learning in the first place. A small clinic joins a federation because it cannot train alone. Our results say that what it receives depends measurably on a choice its coordinator makes before training
starts, and that the choice can be made well for the price of a few minutes of compute.

\subsection{Why Personalization Fails}

Ditto reduced worst-client performance in all three domain-matched cells, and sweeping its regularisation strength over two orders of magnitude changed the outcome by between 0.006 and 0.048. We do not read this as a defect in the method. Ditto performs as designed: it trains a personal model and pulls it toward the global one. The difficulty is what the personal model is fitted on.

At $\alpha = 0.1$ the smallest client holds roughly twenty examples. Regularising a model toward a good one helps only in proportion to how strongly it is pulled, and our sweep shows that proportion is small. The failure is set by client data volume, not by the algorithm.

This suggests a boundary rather than a verdict. Personalization needs a client with enough data to personalize on. Where that condition fails, methods that keep every client on the shared model do better, and our comparison bears this out: FedAvg, FedProx and FFA-LoRA cluster together and stay positive, while Ditto and purely local training fall together well below zero. The dividing line is not between simple and sophisticated methods. It runs between methods that keep the worst client on the shared model and methods that do not.

The explanation offered for worst-client failure in our earlier conference version~\cite{naseer2026when} is interference: the minority client's update conflicts with the aggregate and is overwritten. We measured this directly rather than inferring it from outcomes, and it does not hold. The cosine between the smallest client's update and the aggregate averages
$+0.013$ across all cells. The geometry measured in Section~\ref{sec:res-mechanism} shows no such conflict. Conflict rates are slightly higher in the well-matched cells than in the mismatched ones, which is the reverse of what interference predicts. 

A second explanation is that the frozen representation does the work: a well-matched backbone already separates the client's classes, so few local examples suffice. Our local-only condition tests this directly by removing aggregation. It fails as clearly. In all five cells we ran, the task-specific baseline beat the pretrained model on every single partition.

What remains consistent with both measurements is a weaker and less tidy account. Aggregation supplies the worst client with information it cannot obtain locally, and the backbone governs how efficiently that information can be absorbed. We did not establish this, and we do not claim it. We report it as the hypothesis our
evidence leaves standing, and note what would test it: an intervention that varies how much of the aggregate reaches the smallest client, holding the backbone fixed.

%% file: sections/06_limitations.tex
\section{Limitations and Open Questions}
\label{sec:limitations}

\paragraph{Model scale.}
All three grid backbones are base-scale encoders of 110 to 125 million parameters.
We did not test decoder-only models, instruction-tuned models, or anything above one billion parameters. Whether per-word perplexity orders backbones the same way at that scale is untested. It is a cheap experiment for anyone with the compute.

\paragraph{Number of clients.}
The main grid uses ten cross-silo clients. A matched $K=50$
check on one cell (AG News $\times$ RoBERTa, $n=5$ seeds) shows the backbone-choice effect strengthening rather than weakening at higher client count ($+0.740$ against $+0.124$ at $K=10$, $p=0.0001$): TextCNN's worst client fails outright on two of five seeds, while RoBERTa's holds between 0.79 and 0.84 throughout. This is one cell on one dataset and does not establish the pattern generally. Cross-device federations with hundreds or thousands of participants have a different structure again, and we would not extrapolate the selection rule that far without checking it.

\paragraph{Language and task.}
All three corpora are English single-label classification. We have no evidence about generation, sequence labelling, multilingual settings, or languages with different morphology, where fertility behaves differently and may carry more of the signal than it does here.

\paragraph{One adaptation method.}
We use LoRA at rank 8 on the query and value projections. Ranks 4 and 16 appear in the appendix at five seeds and give somewhat larger effects than rank 8 ($+0.160$ and $+0.164$ against $+0.124$), consistent in sign and order of magnitude; other PEFT families are untested.

\paragraph{Dataset scale.}
Our main grid subsamples every corpus to 20{,}000 training examples. At the full
120{,}000 the benefit fell to about two thirds of its size on both metrics. The
effect survives, but it is clearly a function of how data-starved the smallest
client is, and our headline numbers come from the more starved regime.

\paragraph{Five seeds.}
Five partitions per cell is enough for paired comparisons with bootstrap intervals, and it is not enough for a nonparametric test to reach conventional significance. Individual cells with small effects are consequently underpowered: six of our ten cells have intervals that contain zero. The conclusions we draw rest on the pattern across cells rather than on any single one. The same caution applies to the rank correlation in Section~\ref{sec:res-ppl}: the nine cells are not independent observations, since each of the three backbones and three datasets contributes to three points each, so the reported $p$-value should be read as descriptive rather than as evidence from nine exchangeable draws.

\paragraph{A pre-registered prediction that failed.}
Our pre-registered alignment score did not meet its threshold
($\rho = 0.617$, $p = 0.077$). The predictor we report instead was identified by
exploratory analysis of the same data. We tested it on a backbone held out of that
analysis, with the interval committed in advance, and it held. That is one
out-of-sample test, not a replication, and the rule should be treated as
provisional until it is checked on backbones and tasks we did not touch. The original prediction and the change are both recorded in the deviation log released with the code.

\paragraph{Mechanism.}
We rule out two explanations and offer a third that we did not test. That third account should be read as a direction for future work rather than as a finding.

\paragraph{Form of heterogeneity.}
All experiments use Dirichlet label skew, the standard construction in the literature we build on. It is one form of non-IID data among several; whether backbone perplexity orders worst-client benefit under feature skew or quantity skew is untested.

\paragraph{Perplexity versus general model quality.} 
Perplexity orders backbones by absolute worst-client, tenth-percentile, and mean-client scores, not only by the paired benefit reported in Table~\ref{tab:main} (P10 results in Table~\ref{tab:A4}). This shows the relationship is not an artefact of the TextCNN baseline comparison. It does not show that perplexity predicts worst-client protection specifically: mean-client score is ordered by perplexity just as cleanly, so our nine cells cannot separate a worst-client-specific effect from perplexity tracking overall model quality, of which worst-client protection is one visible consequence.

One methodological point follows from our metric analysis and applies beyond this study. A per-client score computed only on the client's own data is maximised by a constant predictor when that client holds a single class. Any paper reporting
worst-client outcomes under extreme label skew should state which test set its per-client metric uses, because the two choices can order methods differently. In our own results the direction of the Ditto comparison reverses between them.

We would also encourage reporting the tenth percentile alongside the minimum. With ten clients the minimum is a single order statistic and moves considerably with the partition draw. On one of our cells it varies by more than $0.25$ across seeds, which is larger than most of the effects we report.

%% file: sections/07_conclusion.tex
\section{Conclusion}
\label{sec:conclusion}

We asked which design choices protect the worst-off client when a federation
fine-tunes a pretrained model with parameter-efficient methods. Across a
nine-cell grid of backbones and datasets, with five partitions per cell and a
task-specific baseline trained on the identical partition, three answers emerged.

The choice of backbone matters, and it can be settled before training starts.
Ranking candidate backbones by per-word masked-language-model perplexity on the task text reproduced the ranking of their worst-client benefit in all three datasets, with a rank correlation of $-0.867$ across nine cells. We committed an interval in advance for a fourth backbone that had been held out of the analysis, and the measurement fell inside it. Producing the score needs no federated training.

Personalization forfeited most of the benefit for the client it targets. Ditto recovered between 4 and 12 percent of the distance between purely local training and FedAvg, and a hundredfold change in its regularisation strength moved that by $0.006$ to $0.048$ across the three cells. At our most extreme skew the smallest client holds around twenty examples, and pulling it toward a good global model recovers only a small fraction of what federation would have given it. Methods that keep every client on the shared model outperformed those that do not.

The benefit depends on aggregation rather than on the frozen representation alone. With aggregation removed, the task-specific baseline beat the pretrained model on every partition in every cell we tested. We also measured the adapter geometry directly and found no sign of the interference that has been proposed as an explanation: the smallest client's update is close to orthogonal to the aggregate, not opposed to it.

For a practitioner the recommendation is short. Measure per-word perplexity on a
sample of your task text before choosing a backbone, and prefer the lowest. Do not
reach for personalization when your smallest participant has very little data.
Report the worst client and the tenth percentile, not only the mean, because the
mean cannot distinguish a federation that serves everyone from one that leaves a
participant behind.

Our alignment result rests on one out-of-sample test rather than a replication, so
we treat the rule as provisional. Checking it on larger backbones, on generation
tasks and on languages other than English would settle how far it carries. The
code, the run manifest, the committed predictions and the full result log are
released so that this is straightforward to do.

%% file: sections/08_appendix.tex
\section{Full results}
\label{app:full}

Table~\ref{tab:A1} gives the worst-client effect for all nine grid cells under each of the three per-client metrics defined in Section~\ref{sec:metrics}. The main text reports P3; the other two are provided so that any comparison can be recomputed under a different endpoint. Under the non-personalised algorithms every client holds the same global model, so the minimum of P2 across clients equals the global balanced accuracy of that model. P2 therefore measures global competence rather than dispersion, and it varies across clients only under Ditto and local-only training.

\IfFileExists{tables/A1_all_metrics.tex}{\input{tables/A1_all_metrics}}%
{\par\medskip\noindent\textcolor{red}{[A1\_all\_metrics.tex is missing. Run \texttt{make\_tables.py} and upload the tables folder.]}\par\medskip}

Across the same nine cells, the rank correlation between log per-word perplexity and the paired benefit under P10 is $\rho=-0.883$
($p=0.0016$), consistent with the primary result under the minimum
($\rho=-0.867$). Table~\ref{tab:A4} reports the per-cell values.

\IfFileExists{tables/A4_p10.tex}{\input{tables/A4_p10}}%
{\par\medskip\noindent\textcolor{red}{[A4\_p10.tex is missing. Upload it to the tables folder.]}\par\medskip}

Table~\ref{tab:A2} reports every aggregation algorithm on the three
domain-matched cells, and Table~\ref{tab:A3} gives the effect across the
Dirichlet concentration.

\IfFileExists{tables/A2_algorithms.tex}{\input{tables/A2_algorithms}}%
{\par\medskip\noindent\textcolor{red}{[A2\_algorithms.tex is missing. Run \texttt{make\_tables.py} and upload the tables folder.]}\par\medskip}
\IfFileExists{tables/A3_alpha.tex}{\input{tables/A3_alpha}}%
{\par\medskip\noindent\textcolor{red}{[A3\_alpha.tex is missing. Run \texttt{make\_tables.py} and upload the tables folder.]}\par\medskip}

Table~\ref{tab:A5} gives the local-only condition for all five cells tested in Section~\ref{sec:results}, including the two off-diagonal cells not covered by Table~\ref{tab:A2}.

\IfFileExists{tables/A5_localonly.tex}{\input{tables/A5_localonly}}%
{\par\medskip\noindent\textcolor{red}{[A5\_localonly.tex is missing. Upload it to the tables folder.]}\par\medskip}

\section{Ablations and controls}
\label{app:ablations}

Table~\ref{tab:B1} collects the remaining conditions. The number of partitions is
given on every row, because they differ. Conditions with fewer than five
partitions are directional and support no claim in the main text; we report them
so that a reader can see what was run rather than only what was conclusive.

\IfFileExists{tables/B1_ablations.tex}{\input{tables/B1_ablations}}%
{\par\medskip\noindent\textcolor{red}{[B1\_ablations.tex is missing. Run \texttt{make\_tables.py} and upload the tables folder.]}\par\medskip}

Two condition types need comment. The matched-gradient-step rows and the full-size row each use their own baseline, trained under the same condition, so the comparison stays paired. Without that, matching compute for the foundation model alone would reintroduce the confound the control exists to remove.

Table~\ref{tab:B2} reports the client-count check.

\IfFileExists{tables/B2_k50.tex}{\input{tables/B2_k50}}%
{\par\medskip\noindent\textcolor{red}{[B2\_k50.tex is missing. Run \texttt{make\_tables.py} and upload the tables folder.]}\par\medskip}

\section{Protocol deviations}
\label{app:deviations}

The protocol, the metrics, the analysis code and the alignment scores with their signed per-cell predictions were committed before the first federated run. Every departure since is recorded with its date and justification in the deviation log released with the code. The substantive entries are:

\begin{itemize}
\item The grid backbones were changed to a matched-capacity set before any run, so
      that pretraining domain would not be confounded with model size.
      DistilBERT moved to a held-out role.
\item The alignment score was changed from a two-way centred quantity to
      uncentred per-word perplexity after the pre-registered threshold was not
      met. The out-of-sample prediction described in
      Section~\ref{sec:res-prereg} was committed before the corresponding runs
      executed.
\item The primary endpoint was changed from P1 to P3 for the reason given in
      Section~\ref{sec:metrics}. All three metrics were logged from the first run,
      so no experiment was repeated because of this change.
\item Two condition blocks were cancelled when measured compute exceeded the
      available budget. Neither supports a claim in the paper.
\item Analysis excludes result records whose run identifier is absent from the
      final manifest. These are benchmark runs and runs from a superseded
      configuration. They were excluded by provenance, not by inspection of their
      values.
\item The per-dataset training subsample was set to 20{,}000 examples for
      the main grid, rather than the 120{,}000-example Sentiment140 target
      in the frozen protocol. The full-scale condition was retained as a
      single appendix control (Table~\ref{tab:B1}, ``120k training
      examples'') rather than the main-grid design.
\item Communication rounds were set to 30, uniform across $\alpha$, rather
      than the 60 rounds specified in the frozen protocol, when measured
      compute exceeded the available budget.
\item The Dirichlet sweep was reduced from $\alpha \in \{0.05, 0.1, 0.3,
      0.5, 1.0\}$ on five cells to $\alpha \in \{0.1, 0.3, 1.0\}$ on the
      three domain-matched cells (Table~\ref{tab:A3}), for the same reason.
\item ``Two condition blocks were cancelled'' above refers specifically to
      the Phase-2 method candidates (\texttt{ours\_a/b/c}) and the
      10-seed method-vs-Ditto endpoint; neither is reported in this paper.
\item The $K=50$ baseline and the LoRA rank 4/16 conditions were completed to
      five seeds after initial submission preparation, closing gaps flagged during internal review. The $K=50$ check shows the backbone effect strengthening at higher client count ($+0.740$ vs $+0.124$ at $K=10$); rank 4/16 give effects of similar order to rank 8. Full values in Tables~\ref{tab:B1} and~\ref{tab:B2}. No claim in the main text was altered.
\end{itemize}

\section{Computational details}
\label{app:compute}

All experiments ran on a single NVIDIA Tesla T4. A foundation-model run of 30 rounds with ten clients on 20{,}000 training examples takes approximately 0.9 GPU-hours; the TextCNN baseline takes about one seventh of that. The complete
study is 313 runs with no failures.

Every run is a row in a manifest keyed by a hash of its full configuration, which
makes duplicate work impossible and lets an interrupted session resume from a
per-round checkpoint. Results are appended to a single log as one record per
evaluated round, never overwritten. We release the code, the manifest, the
committed alignment table, the deviation log and the complete result log, together
with the scripts that regenerate every table and figure in this paper directly
from that log.

%% file: tables/A1_all_metrics.tex
\begin{table}[t]
\centering
\caption{Worst-client effect against the TextCNN baseline at $\alpha = 0.1$ under all three per-client metrics. Five partitions per cell, paired differences.}
\label{tab:A1}
\small
\begin{tabular}{llrrrrrr}
\toprule
Dataset & Backbone & \multicolumn{2}{c}{P3} & \multicolumn{2}{c}{P1} & \multicolumn{2}{c}{P2} \\
\cmidrule(lr){3-4}\cmidrule(lr){5-6}\cmidrule(lr){7-8}
 & & Effect & $p$ & Effect & $p$ & Effect & $p$ \\
\midrule
AG News & RoBERTa & +0.124 & 0.013 & +0.106 & 0.022 & +0.089 & 0.013 \\
AG News & BERTweet & +0.111 & 0.018 & +0.081 & 0.014 & +0.072 & 0.001 \\
AG News & PubMedBERT & +0.017 & 0.657 & $-$0.031 & 0.654 & +0.025 & 0.121 \\
Sentiment140 & RoBERTa & +0.074 & 0.291 & +0.049 & 0.561 & +0.089 & 0.062 \\
Sentiment140 & BERTweet & +0.049 & 0.651 & +0.003 & 0.973 & +0.093 & 0.106 \\
Sentiment140 & PubMedBERT & +0.015 & 0.568 & +0.037 & 0.373 & +0.020 & 0.394 \\
PubMed & RoBERTa & +0.140 & 0.004 & +0.104 & 0.081 & +0.111 & 0.001 \\
PubMed & BERTweet & $-$0.019 & 0.357 & $-$0.033 & 0.551 & +0.015 & 0.496 \\
PubMed & PubMedBERT & +0.208 & 0.004 & +0.171 & 0.029 & +0.165 $<0.001$\\
\bottomrule
\end{tabular}
\end{table}

%% file: tables/A4_p10.tex
\begin{table}[h]
\centering
\caption{Worst-client benefit under the tenth percentile (P10) of
client-class macro-F1, same nine grid cells and dataset ordering as
Table~4. Paired difference against TextCNN, $\alpha=0.1$, five seeds.}
\label{tab:A4}
\small
\begin{tabular}{llrr}
\toprule
Dataset & Backbone & Effect & $p$ \\
\midrule
AG News & RoBERTa & +0.112 & 0.0224 \\
AG News & BERTweet & +0.096 & 0.0043 \\
AG News & PubMedBERT & +0.024 & 0.4114 \\
Sentiment140 & RoBERTa & +0.046 & 0.5892 \\
Sentiment140 & BERTweet & +0.022 & 0.8614 \\
Sentiment140 & PubMedBERT & -0.009 & 0.8351 \\
PubMed & PubMedBERT & +0.189 & 0.0030 \\
PubMed & RoBERTa & +0.121 & 0.0063 \\
PubMed & BERTweet & -0.010 & 0.6673 \\
\bottomrule
\end{tabular}
\end{table}

%% file: tables/A2_algorithms.tex
\begin{table}[t]
\centering
\caption{Worst-client effect (P3) by aggregation algorithm on the three domain-matched cells, $\alpha = 0.1$, five partitions each.}
\label{tab:A2}
\small
\begin{tabular}{lrrr}
\toprule
Algorithm & AG News & Sentiment140 & PubMed \\
\midrule
FedAvg & +0.124 & +0.049 & +0.208 \\
FedProx & +0.113 & +0.058 & +0.206 \\
FFA-LoRA & +0.117 & +0.100 & +0.144 \\
Ditto & $-$0.520 & $-$0.490 & $-$0.203 \\
Local only & $-$0.584 & $-$0.476 & $-$0.221 \\
\bottomrule
\end{tabular}
\end{table}

%% file: tables/A3_alpha.tex
\begin{table}[t]
\centering
\caption{Worst-client effect (P3) across the Dirichlet concentration.}
\label{tab:A3}
\small
\begin{tabular}{lrrr}
\toprule
Cell & $\alpha=0.1$ & $\alpha=0.3$ & $\alpha=1.0$ \\
\midrule
AG News $\times$ RoBERTa & +0.124 & +0.085 & +0.044 \\
Sentiment140 $\times$ BERTweet & +0.049 & +0.142 & +0.122 \\
PubMed $\times$ PubMedBERT & +0.208 & +0.139 & +0.101 \\
\bottomrule
\end{tabular}
\end{table}

%% file: tables/A5_localonly.tex
\begin{table}[h]
\centering
\caption{Worst-client effect (P3) with aggregation removed, all five cells
tested. Local-only training against the identical partition, five seeds.}
\label{tab:A5}
\small
\begin{tabular}{lrr}
\toprule
Cell & Effect & Cliff's $\delta$ \\
\midrule
AG News $\times$ RoBERTa & $-0.584$ & $-1.00$ \\
Sentiment140 $\times$ BERTweet & $-0.476$ & $-1.00$ \\
PubMed $\times$ PubMedBERT & $-0.221$ & $-1.00$ \\
PubMed $\times$ BERTweet & $-0.249$ & $-1.00$ \\
Sentiment140 $\times$ PubMedBERT & $-0.414$ & $-1.00$ \\
\bottomrule
\end{tabular}
\end{table}

%% file: tables/B1_ablations.tex
\begin{table}[t]
\centering
\caption{Ablations and controls. Conditions with fewer than five partitions are directional and are not used to support any claim in the main text. The main-grid reference for AG News $\times$ RoBERTa is given in the final row; the references for the other two matched-gradient-step cells are the FedAvg values in Table~\ref{tab:A2}.}
\label{tab:B1}
\small
\begin{tabular}{llrr}
\toprule
Condition & Cell & Effect & $n$ seeds \\
\midrule
LoRA rank 4 & AG News $\times$ RoBERTa & +0.160 & 5 \\
LoRA rank 16 & AG News $\times$ RoBERTa & +0.164 & 5 \\
DistilBERT (held out) & AG News $\times$ DistilBERT & +0.091 & 5 \\
FedAvgW $\beta=0.1$ & AG News $\times$ DistilBERT & +0.237 & 2 \\
FedAvgW $\beta=0.5$ & AG News $\times$ DistilBERT & +0.174 & 2 \\
Matched gradient steps & AG News $\times$ RoBERTa & +0.338 & 5 \\
Matched gradient steps & Sentiment140 $\times$ BERTweet & +0.121 & 5 \\
Matched gradient steps & PubMed $\times$ PubMedBERT & +0.210 & 5 \\
120k training examples & AG News $\times$ RoBERTa & +0.083 & 5 \\
\midrule
Reference (main grid) & AG News $\times$ RoBERTa & +0.124 & 5 \\
\bottomrule
\end{tabular}
\end{table}

%% file: tables/B2_k50.tex
\begin{table}[h]
\centering
\caption{Client-count check, paired against the matched-baseline design used throughout this paper.}
\label{tab:B2}
\begin{tabular}{lccc}
\toprule
Condition & Worst-client P3 effect & $p$ & $n$ seeds \\
\midrule
$K=10$ (main grid) & $+0.124$ & 0.013 & 5 \\
$K=50$             & $+0.740$ & 0.0001 & 5 \\
\bottomrule
\end{tabular}
\end{table}

%% file: main.bbl
\begin{thebibliography}{999}

\bibitem[Nguyen et~al.(2023)Nguyen, Wang, Malik, Sanjabi, and
  Rabbat]{nguyen2023where}
Nguyen, J.; Wang, J.; Malik, K.; Sanjabi, M.; Rabbat, M.
\newblock Where to Begin? On the Impact of Pre-Training and Initialization in
  Federated Learning.
\newblock In Proceedings of the International Conference on Learning
  Representations (ICLR),  2023.
\newblock arXiv:2206.15387.

\bibitem[Li et~al.(2021)Li, Hu, Beirami, and Smith]{li2021ditto}
Li, T.; Hu, S.; Beirami, A.; Smith, V.
\newblock Ditto: Fair and Robust Federated Learning Through Personalization.
\newblock In Proceedings of the Proceedings of the 38th International
  Conference on Machine Learning (ICML),  2021, Vol. 139, {\em PMLR}, pp.
  6357--6368.

\bibitem[McMahan et~al.(2017)McMahan, Moore, Ramage, Hampson, and Ag{\"u}era~y
  Arcas]{mcmahan2017fedavg}
McMahan, B.; Moore, E.; Ramage, D.; Hampson, S.; Ag{\"u}era~y Arcas, B.
\newblock Communication-Efficient Learning of Deep Networks from Decentralized
  Data.
\newblock In Proceedings of the Proceedings of the 20th International
  Conference on Artificial Intelligence and Statistics (AISTATS),  2017, pp.
  1273--1282.

\bibitem[Li et~al.(2020)Li, Sahu, Zaheer, Sanjabi, Talwalkar, and
  Smith]{li2020fedprox}
Li, T.; Sahu, A.K.; Zaheer, M.; Sanjabi, M.; Talwalkar, A.; Smith, V.
\newblock Federated Optimization in Heterogeneous Networks.
\newblock In Proceedings of the Proceedings of Machine Learning and Systems
  (MLSys),  2020, Vol.~2, pp. 429--450.

\bibitem[Karimireddy et~al.(2020)Karimireddy, Kale, Mohri, Reddi, Stich, and
  Suresh]{karimireddy2020scaffold}
Karimireddy, S.P.; Kale, S.; Mohri, M.; Reddi, S.J.; Stich, S.U.; Suresh, A.T.
\newblock {SCAFFOLD}: Stochastic Controlled Averaging for Federated Learning.
\newblock In Proceedings of the Proceedings of the 37th International
  Conference on Machine Learning (ICML),  2020, pp. 5132--5143.

\bibitem[Hsu et~al.(2019)Hsu, Qi, and Brown]{hsu2019measuring}
Hsu, T.M.H.; Qi, H.; Brown, M.
\newblock Measuring the Effects of Non-Identical Data Distribution for
  Federated Visual Classification.
\newblock {\em arXiv preprint arXiv:1909.06335} {\bf 2019}.

\bibitem[Jimenez~Gutierrez et~al.(2024)Jimenez~Gutierrez, Solans, Heikkil{\"a},
  Vitaletti, Kourtellis, Anagnostopoulos, and
  Chatzigiannakis]{jimenez2024noniid}
Jimenez~Gutierrez, D.M.; Solans, D.; Heikkil{\"a}, M.A.; Vitaletti, A.;
  Kourtellis, N.; Anagnostopoulos, A.; Chatzigiannakis, I.
\newblock Non-{IID} Data in Federated Learning: A Survey with Taxonomy,
  Metrics, Methods, Frameworks and Future Directions.
\newblock {\em arXiv preprint arXiv:2411.12377} {\bf 2024}.

\bibitem[Hu et~al.(2022)Hu, Shen, Wallis, Allen-Zhu, Li, Wang, Wang, and
  Chen]{hu2022lora}
Hu, E.J.; Shen, Y.; Wallis, P.; Allen-Zhu, Z.; Li, Y.; Wang, S.; Wang, L.;
  Chen, W.
\newblock {LoRA}: Low-Rank Adaptation of Large Language Models.
\newblock In Proceedings of the International Conference on Learning
  Representations (ICLR),  2022.

\bibitem[Sun et~al.(2024)Sun, Li, Li, and Ding]{sun2024ffalora}
Sun, Y.; Li, Z.; Li, Y.; Ding, B.
\newblock Improving {LoRA} in Privacy-preserving Federated Learning.
\newblock In Proceedings of the International Conference on Learning
  Representations (ICLR),  2024.
\newblock arXiv:2403.12313.

\bibitem[Guo et~al.(2025)Guo, Zeng, Wang, Fan, Wang, and Qu]{guo2025fedsalora}
Guo, P.; Zeng, S.; Wang, Y.; Fan, H.; Wang, F.; Qu, L.
\newblock Selective Aggregation for Low-Rank Adaptation in Federated Learning.
\newblock In Proceedings of the International Conference on Learning
  Representations (ICLR),  2025.
\newblock arXiv:2410.01463.

\bibitem[Wang et~al.(2024)Wang, Shen, He, Sun, Wang, Lyu, and
  Li]{wang2024flora}
Wang, Z.; Shen, Z.; He, Y.; Sun, G.; Wang, H.; Lyu, L.; Li, A.
\newblock {FLoRA}: Federated Fine-Tuning Large Language Models with
  Heterogeneous Low-Rank Adaptations.
\newblock In Proceedings of the Advances in Neural Information Processing
  Systems (NeurIPS),  2024, Vol.~37, pp. 22513--22533.

\bibitem[Cho et~al.(2024)Cho, Liu, Xu, Fahrezi, and
  Joshi]{cho2024heterogeneous}
Cho, Y.J.; Liu, L.; Xu, Z.; Fahrezi, A.; Joshi, G.
\newblock Heterogeneous {LoRA} for Federated Fine-tuning of On-Device
  Foundation Models.
\newblock In Proceedings of the Proceedings of the 2024 Conference on Empirical
  Methods in Natural Language Processing (EMNLP),  2024, pp. 12903--12913.

\bibitem[Bai et~al.(2024)Bai, Chen, Qian, Yao, and Li]{bai2024flexlora}
Bai, J.; Chen, D.; Qian, B.; Yao, L.; Li, Y.
\newblock Federated Fine-tuning of Large Language Models under Heterogeneous
  Tasks and Client Resources.
\newblock In Proceedings of the Advances in Neural Information Processing
  Systems (NeurIPS),  2024.

\bibitem[Zhang et~al.(2024)Zhang, Vahidian, Kuo, Li, Zhang, Yu, Wang, and
  Chen]{zhang2024fedit}
Zhang, J.; Vahidian, S.; Kuo, M.; Li, C.; Zhang, R.; Yu, T.; Wang, G.; Chen, Y.
\newblock Towards Building the Federated {GPT}: Federated Instruction Tuning.
\newblock In Proceedings of the IEEE International Conference on Acoustics,
  Speech and Signal Processing (ICASSP),  2024, pp. 6915--6919.

\bibitem[Yang et~al.(2025)Yang, Long, Lu, Zhu, Jiang, and
  Zhang]{yang2025fedlorasurvey}
Yang, Y.; Long, G.; Lu, Q.; Zhu, L.; Jiang, J.; Zhang, C.
\newblock Federated Low-Rank Adaptation for Foundation Models: A Survey.
\newblock {\em arXiv preprint arXiv:2505.13502} {\bf 2025}.

\bibitem[Mohri et~al.(2019)Mohri, Sivek, and Suresh]{mohri2019agnostic}
Mohri, M.; Sivek, G.; Suresh, A.T.
\newblock Agnostic Federated Learning.
\newblock In Proceedings of the Proceedings of the 36th International
  Conference on Machine Learning (ICML),  2019, Vol.~97, {\em PMLR}, pp.
  4615--4625.

\bibitem[Li et~al.(2020)Li, Sanjabi, Beirami, and Smith]{li2020qfedavg}
Li, T.; Sanjabi, M.; Beirami, A.; Smith, V.
\newblock Fair Resource Allocation in Federated Learning.
\newblock In Proceedings of the International Conference on Learning
  Representations (ICLR),  2020.

\bibitem[Fallah et~al.(2020)Fallah, Mokhtari, and
  Ozdaglar]{fallah2020perfedavg}
Fallah, A.; Mokhtari, A.; Ozdaglar, A.
\newblock Personalized Federated Learning with Theoretical Guarantees: A
  Model-Agnostic Meta-Learning Approach.
\newblock In Proceedings of the Advances in Neural Information Processing
  Systems (NeurIPS),  2020, Vol.~33, pp. 3557--3568.

\bibitem[T.~Dinh et~al.(2020)T.~Dinh, Tran, and Nguyen]{dinh2020pfedme}
T.~Dinh, C.; Tran, N.H.; Nguyen, T.D.
\newblock Personalized Federated Learning with {M}oreau Envelopes.
\newblock In Proceedings of the Advances in Neural Information Processing
  Systems (NeurIPS),  2020, Vol.~33, pp. 21394--21405.

\bibitem[Yi et~al.(2023)Yi, Yu, Wang, Liu, and Li]{yi2023pfedlora}
Yi, L.; Yu, H.; Wang, G.; Liu, X.; Li, X.
\newblock {pFedLoRA}: Model-Heterogeneous Personalized Federated Learning with
  {LoRA} Tuning.
\newblock {\em arXiv preprint arXiv:2310.13283} {\bf 2023}.

\bibitem[Salazar et~al.(2024)Salazar, Ara{\'u}jo, Cano, and
  Abreu]{salazar2024groupfairness}
Salazar, T.; Ara{\'u}jo, H.; Cano, A.; Abreu, P.H.
\newblock A Survey on Group Fairness in Federated Learning: Challenges,
  Taxonomy of Solutions and Directions for Future Research.
\newblock {\em arXiv preprint arXiv:2410.03855} {\bf 2024}.

\bibitem[Heilmann et~al.(2025)Heilmann, Corbucci, and
  Cerrato]{heilmann2025benchmark}
Heilmann, X.; Corbucci, L.; Cerrato, M.
\newblock A Benchmark for Client-level Fairness in Federated Learning.
\newblock In Proceedings of the Proceedings of the Fourth European Workshop on
  Algorithmic Fairness,  2025, Vol. 294, {\em PMLR}, pp. 434--438.

\bibitem[Naseer and Shoaib(2026)]{naseer2026when}
Naseer, K.; Shoaib, U.
\newblock When More Parameters Hurt: Foundation Model Priors Amplify
  Worst-Client Disparity Under Extreme Federated Heterogeneity.
\newblock {\em arXiv preprint arXiv:2605.08992} {\bf 2026}.

\bibitem[Liu et~al.(2019)Liu, Ott, Goyal, Du, Joshi, Chen, Levy, Lewis,
  Zettlemoyer, and Stoyanov]{liu2019roberta}
Liu, Y.; Ott, M.; Goyal, N.; Du, J.; Joshi, M.; Chen, D.; Levy, O.; Lewis, M.;
  Zettlemoyer, L.; Stoyanov, V.
\newblock {RoBERTa}: A Robustly Optimized {BERT} Pretraining Approach.
\newblock {\em arXiv preprint arXiv:1907.11692} {\bf 2019}.

\bibitem[Nguyen et~al.(2020)Nguyen, Vu, and Nguyen]{nguyen2020bertweet}
Nguyen, D.Q.; Vu, T.; Nguyen, A.T.
\newblock {BERTweet}: A pre-trained language model for {E}nglish Tweets.
\newblock In Proceedings of the Proceedings of the 2020 Conference on Empirical
  Methods in Natural Language Processing: System Demonstrations,  2020, pp.
  9--14.

\bibitem[Gu et~al.(2021)Gu, Tinn, Cheng, Lucas, Usuyama, Liu, Naumann, Gao, and
  Poon]{gu2021pubmedbert}
Gu, Y.; Tinn, R.; Cheng, H.; Lucas, M.; Usuyama, N.; Liu, X.; Naumann, T.; Gao,
  J.; Poon, H.
\newblock Domain-Specific Language Model Pretraining for Biomedical Natural
  Language Processing.
\newblock {\em ACM Transactions on Computing for Healthcare} {\bf 2021}, {\em
  3},~1--23.

\bibitem[Kim(2014)]{kim2014textcnn}
Kim, Y.
\newblock Convolutional Neural Networks for Sentence Classification.
\newblock In Proceedings of the Proceedings of the 2014 Conference on Empirical
  Methods in Natural Language Processing (EMNLP),  2014, pp. 1746--1751.

\bibitem[Zhang et~al.(2015)Zhang, Zhao, and LeCun]{zhang2015character}
Zhang, X.; Zhao, J.; LeCun, Y.
\newblock Character-level Convolutional Networks for Text Classification.
\newblock In Proceedings of the Advances in Neural Information Processing
  Systems (NeurIPS),  2015, Vol.~28, pp. 649--657.

\bibitem[Go et~al.(2009)Go, Bhayani, and Huang]{go2009twitter}
Go, A.; Bhayani, R.; Huang, L.
\newblock Twitter Sentiment Classification using Distant Supervision.
\newblock Technical report, Stanford University,  2009.
\newblock CS224N Project Report.

\bibitem[Dernoncourt and Lee(2017)]{dernoncourt2017pubmed}
Dernoncourt, F.; Lee, J.Y.
\newblock {PubMed} 200k {RCT}: a Dataset for Sequential Sentence Classification
  in Medical Abstracts.
\newblock In Proceedings of the Proceedings of the Eighth International Joint
  Conference on Natural Language Processing (IJCNLP),  2017, pp. 308--313.

\bibitem[Salazar et~al.(2020)Salazar, Liang, Nguyen, and
  Kirchhoff]{salazar2020masked}
Salazar, J.; Liang, D.; Nguyen, T.Q.; Kirchhoff, K.
\newblock Masked Language Model Scoring.
\newblock In Proceedings of the Proceedings of the 58th Annual Meeting of the
  Association for Computational Linguistics (ACL),  2020, pp. 2699--2712.

\bibitem[Rust et~al.(2021)Rust, Pfeiffer, Vuli{\'c}, Ruder, and
  Gurevych]{rust2021good}
Rust, P.; Pfeiffer, J.; Vuli{\'c}, I.; Ruder, S.; Gurevych, I.
\newblock How Good is Your Tokenizer? On the Monolingual Performance of
  Multilingual Language Models.
\newblock In Proceedings of the Proceedings of the 59th Annual Meeting of the
  Association for Computational Linguistics (ACL),  2021, pp. 3118--3135.

\end{thebibliography}
